\documentclass[conference]{IEEEtran}
\ifCLASSINFOpdf
  \usepackage[pdftex]{graphicx}
  \graphicspath{{.}}
\else
  \usepackage[dvips]{graphicx}
  \graphicspath{{.}}
\fi
\begin{document}
%
\title{Solving a Path Planning Problem in a Partially Known Environment using a Swarm Algorithm}

\author{\IEEEauthorblockN{Esh Vckay ({\em Student Trainee}) ~~Mansimar Aneja ({\em Student Trainee}) ~Dipti Deodhare}
\IEEEauthorblockA{Centre for Artificial Intelligence and Robotics\\
DRDO, Ministry of Defence\\
Kaggadasapura Main Road, Bangalore - 560 003.\\
Email: esh@eshvk.me, mansimaraneja@gmail.com, dipti@cair.drdo.in}}


%


\maketitle

\begin{abstract}
This paper proposes a path planning strategy for an Autonomous Ground Vehicle (AGV) navigating in a partially known environment. Global path planning is performed by first using a spatial database of the region to be traversed containing selected attributes such as height data and soil information from a suitable spatial database. The database is processed using a biomimetic swarm algorithm that is inspired by the nest building strategies followed by termites. Local path planning is performed online utilizing information regarding contingencies that affect the safe navigation of the AGV from various sensors. The simulation discussed has been implemented on the open source {\it Player-Stage-Gazebo} platform. 
\end{abstract}
\vspace{0.1in}
\noindent {\bf Keywords:} Swarm Intelligence, Distributed algorithms, Path Planning

%
\IEEEpeerreviewmaketitle

\section{Introduction}
\label{sec:intro}
The design of a complete planner that computes a collision free path for a non holonomic robot in any partially known environment especially outdoor terrains is constrained by a number of factors. These include the number of degrees of freedom imposed by the robot's geometry, the presence or absence of obstacles and sensor inaccuracies~\cite{1Laumond} which cause an exponential increase in computational complexity of the problem~\cite{2Canny}. One strategy to reduce the complexity of the problem is to formulate assumptions that simplify the environment~\cite{3Stappen,4Stappen,5Schwartz}. Another strategy is to design planners that satisfy weaker forms of completeness such as the probablistic path planner~\cite{6Svetska} and the randomized path planner~\cite{7Barraquand}. An auxiliary strategy presented here is to reduce the computational processing requirments by the use of a distributed algorithm based on the principle of swarm intelligence. Swarm intelligence involves the design of algorithms based on the interactions between social insects such as termites and ants which leads to emergent intelligent behaviour~\cite{8Bonabeau}.

Approaches to designing planners for partially or dynamic environments differ in the way the environment is modelled. Topological maps utilize graphs where nodes represent landmark information and edges represent connections between them~\cite{9Heero}. Planners based on topological maps include~\cite{10Haigh}. The main problem of utilizing such approaches in outdoor terrains is that of misidentification of similar landmarks. Metric maps capture geometric properties of the environment. Planners utilizing metric information include the D* algorithm~\cite{11Stentz,12Stentz}, potential field planners~\cite{13Choset} and wave-front based planners~\cite{14Hughes,15Jahanbin}. For large areas, although metric maps provide finer resolutions allowing for more detailed planning to take place, processing requirements increase drastically.

\section{The Swarm Algorithm}
\label{sec:algo}
\begin{figure}[h]
\centering
\includegraphics[width=2in]{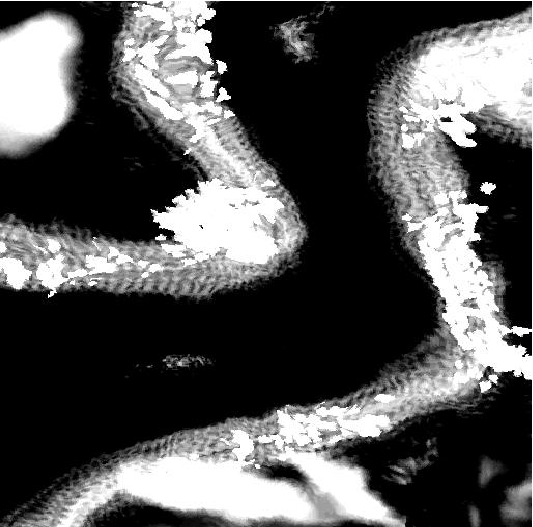}
\caption{Grey scale representation of the height map.}
\label{fig:Turbulence}
\end{figure}

\begin{figure}[h]
\centering
\includegraphics[width=2in]{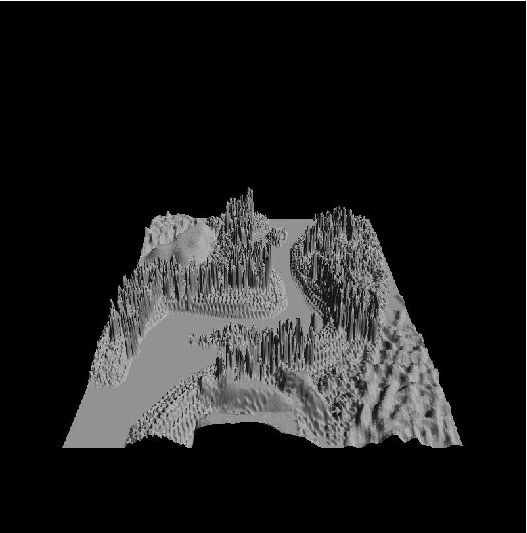}
\caption{The height map rendered in 3D.}
\label{fig:3DHeight}
\end{figure}

The generation of a desired collision free path involves two steps namely, ({\em i}) Global Path planning and ({\em ii}) Local Path planning. Here, global path planning is performed by first generating a spatial database of the region with selected features such as soil and gradient information. Gradient information is generated from  Digital Elevation Models (DEM) which  are digital representations of ground surface topography~\cite{16Wilson}. Elevation data having 30-90 metre resolution is freely available from the Shuttle Radar Topography Mission (SRTM)~\cite{17Rabus}. 
Other sources of DEMs include~\cite{18usgs}. Figure~\ref{fig:Turbulence} depicts the grey scale height map which is used as an input to generate the terrain. The corresponding terrain rendered in 3D is shown in Figure~\ref{fig:3DHeight}. Every pixel in a height map has a value varying from 0-255. 
A black spot corresponds to minimum elevation and a white spot corresponds to maximum elevation. In our simulation, this data is stored in a Postgres database table. This is mainly done to facilitate integration with soil information. To optimize space requirements, the data is sub-sampled so that every cell in the table is an average of the height map values of a group of 8 pixels. 
This height information is used to compute the gradient at each point/cell, in different directions and is simply the difference in the height values between two neighbouring points. The Pioneer2DX robot can travel safely on terrin bounded by a 25\% limit on grade (an angle of approximately 15 deg. with the horizontal). Once the gradient information is computed, the data is mapped to a range of values ranging from 1 to 9 as shown in Table~\ref{tbl:gradvals} below. 

\begin{table}[!h]
\centering
\caption{Generation of Gradient-based Rank Values from Height Difference}
\begin{small}
\begin{tabular}{||c|c||} \hline
Cell Height Difference & Gradient Goodness Value \\ \hline \hline
255~-~193 & 1 \\ \hline
192~-~130 & 2 \\ \hline
126~-~67 & 3 \\ \hline
66~-~1 & 4 \\ \hline
0 & 5 \\ \hline
-1~-~-66 & 4 \\ \hline
-67~-~-129 & 3 \\ \hline
-130~-~-192 & 2 \\ \hline
-192 ~- ~-255 & 1 \\ \hline
\end{tabular}
\end{small}
\label{tbl:gradvals}
\end{table}

Soil information of the region is obtained and classified into five categories that place bounds on robot manouverability. Soil information of the region is obtained from the GRASS POSTGIS database in the form of a vector map where the various soil types are classified by a parameter called {\em cat value}. This information is parsed and classified into five different categories namely, {\em Gravel}, {\em Sand}, {\em Clay}, {\em Silt} and {\em Rock}, in order of suitability for vehicle traversal. To facilitate computation these categories are assigned values 5, 4, 3, 2, 1, respectively to indicate the ``goodness'' for traversability. Table~\ref{tbl:soilgoodness} records this map formally.

\begin{table}[!h]
\centering
\caption{Generation of Soil Goodness Values}
\begin{small}
\begin{tabular}{||c|c||} \hline
Soil Category & Soil Goodness Value \\ \hline \hline
Gravel & 5 \\ \hline
Sand & 4 \\ \hline
Clay & 3 \\ \hline
Silt & 3 \\ \hline
Rock & 1 \\ \hline
\end{tabular}
\end{small}
\label{tbl:soilgoodness}
\end{table}

Once the destination and start points are selected on the map, an arc joining the two is generated and a wide swathe of grid cells around the region are selected for processing. A biomimetic swarm algorithm inspired by nest building strategies in termites is used to process the data. The algorithm has already been proven successful in optimizing site selection in Geographical Information Systems (GIS)~\cite{19Sharma}. A rank map is first generated  where a cell is defined to be good or bad by generating a single rank value by adding the Gradient Goodness Values and Soil Goodness Values discussed above. The expression for computing rank is given as equation~\ref{eq:rank}.
\begin{equation}
\label{eq:rank}
{\rm Rank}, ~R = {\rm Soil Goodness Value} + {\rm Height Goodness Value}
\end{equation}
This is a very simple way of computing the rank. More formal and complex Multi-Criteria Decision Making (MCDM) strategies like the Analytical Hierarchical Process (AHP) can be used. Once the rank has been obtained we now use a swarm algorithm inspired by the nest building behaviour of termites to perform the global path planning.

\begin{figure}[h]
\centering
\includegraphics[width=2.5in]{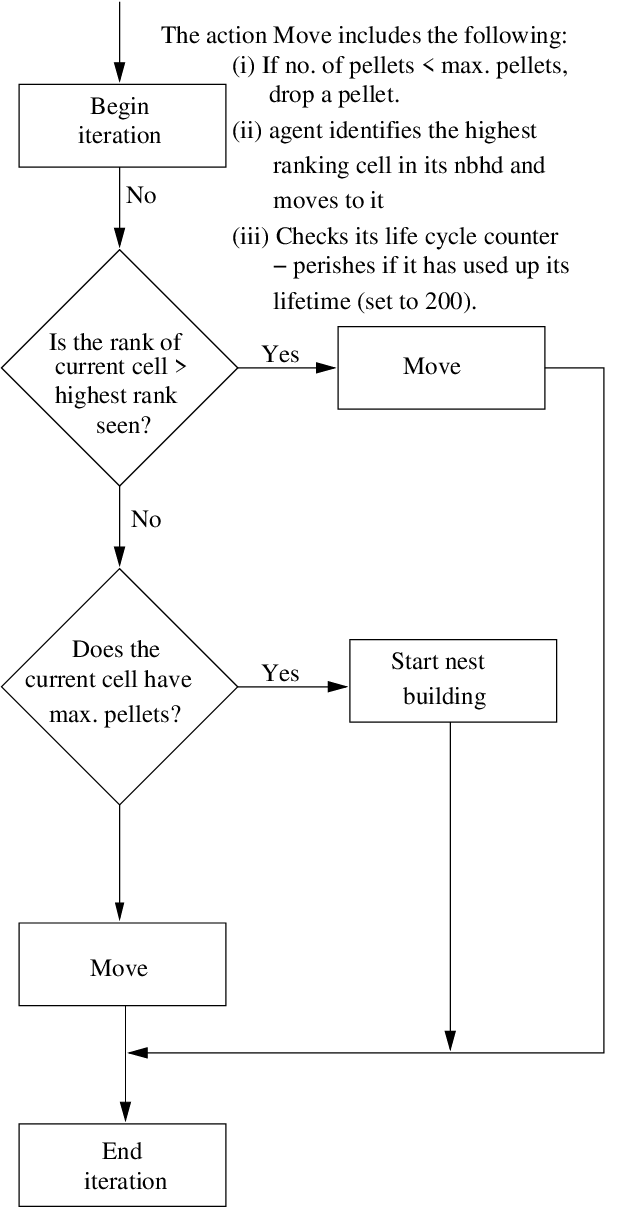}
\caption{Flowchart of the Swarm Algorithm}
\label{fig:swarm1}
\end{figure}

\begin{figure}[h]
\centering
\includegraphics[width=2.5in]{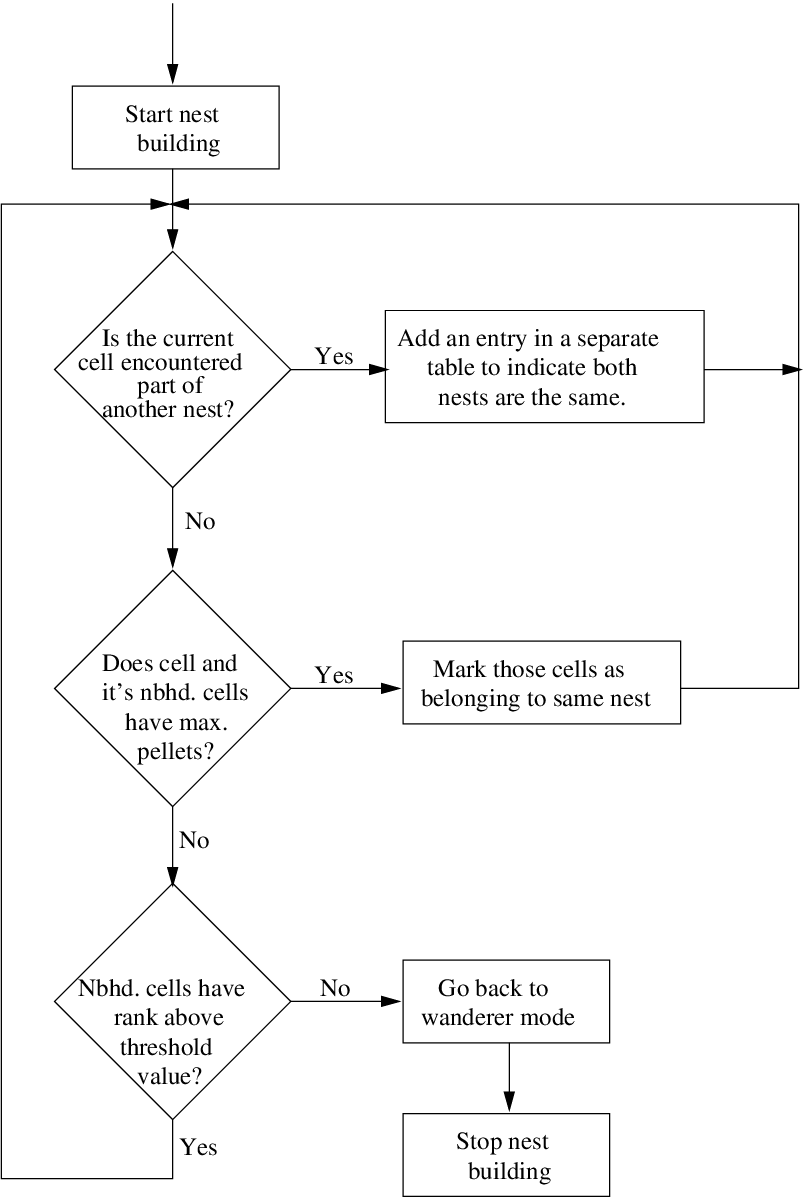}
\caption{Flowchart of the Nest Building Subroutine}
\label{fig:swarm2}
\end{figure}

\begin{figure}[h]
\centering
\includegraphics[width=2.25in]{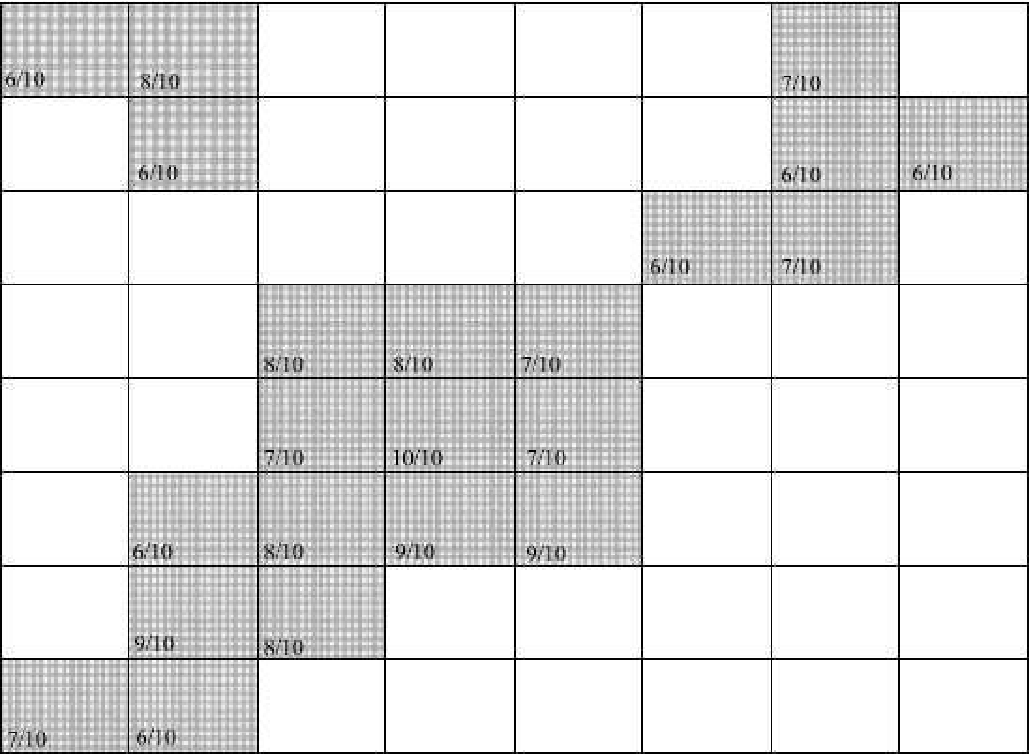}
\caption{Grid for Path Planning.}
\label{fig:gridlayout}
\end{figure}

The algorithm is summarized as follows: A swarm of agents endowed with simplistic rules that govern their behaviour and local interaction is deployed randomly on the map to determine those cells that satisfy the constraints imposed by the vehicle for safe navigation. Each agent's interaction rules lead to two behaviours, namely, {\em pellet dropping} and {\em nest building}. In the pellet dropping stage, agents utilize permanent pheromones in the form of pellets to mark cells of a suitable rank. Once the number of pellets reaches a certain maximum limit, agents use the cell as a focal point to move to the next stage, namely nest building. In the nest building mode, agents forage in the local neighborhood of the suitable cell looking for other cells which satisfy the navigation criteria. If all the cells in the local vicinity of the cell with maximum number of pellets satisfy the navigation criteria, a nest of these cells is then created. When two nests created by different agents come in contact, they are merged together to create large contiguous areas suitable for navigation. Figure~\ref{fig:gridlayout} illustrates the rank map, where the cells are ranked and declared as part of a nest, as indicated by the shaded patterns. The flow chart of the swarm algorithm is included in Figures~\ref{fig:swarm1} and~\ref{fig:swarm2}.

\begin{figure}[h]
\centering
\includegraphics[width=2.25in]{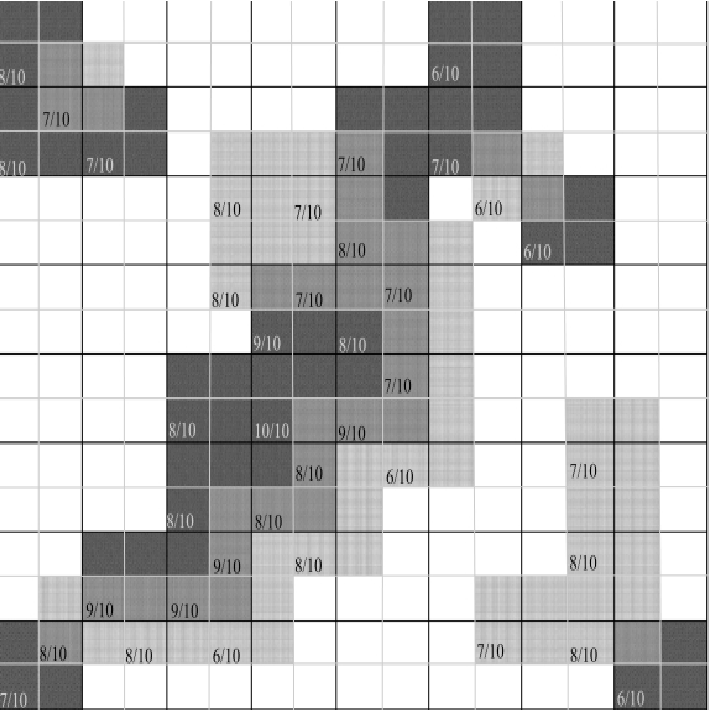}
\caption{Two Overlapping Grids.}
\label{fig:gridongrid}
\end{figure}

Grid maps have a major limitation in that the path produced can become suboptimal due to unnecessary increase in length. This is caused by wastage in space as cells are marked occupied by obstacles even if only a small portion of the cell is occupied by an obstacle. One possible solution is to increase the grid resolution which however leads to increase in processing requirements. Such a restriction is partially removed by generating two grid maps where each cell is four times the Robot size and the maps are separated by an offset which is equivalent to one fourth of the cell size. The  resultant map consists of  overlapping grid cells that are suitably ranked and nested, as can be seen in Figure~\ref{fig:gridongrid}. Both maps are continously checked during the robot's navigation.

\begin{figure}[h]
\centering
\includegraphics[width=2in]{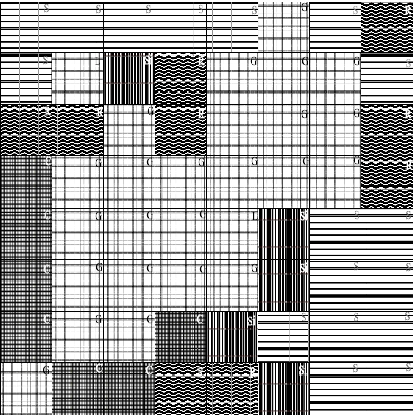}
\caption{Grid indicating different soil categories.}
\label{fig:soiltypes1}
\end{figure}

\begin{figure}[h]
\centering
\includegraphics[width=2in]{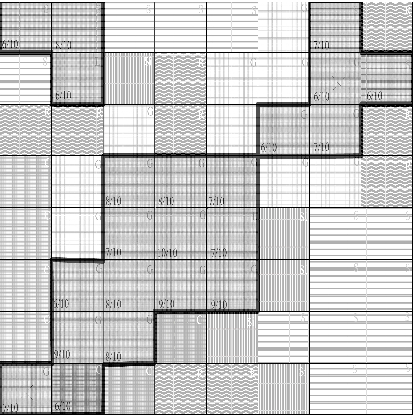}
\caption{Ranked cells indicating both soil and height data combined. Outlined regions are nests (navigable regions) detected by the swarm algorithm.}
\label{fig:soiltypes2}
\end{figure}

\section{Local Path Planning}
In a partially known environment, real time obstacle avoidance is performed by utilizing sensory information  regarding contingencies that affect the safe navigation of the robot. A key factor in designing a obstacle avoidance algorithm is minimising the generation of non-optimal local paths. Various obstacle avoidance techniques include the edge detection~\cite{20Kuc} and certainty grid technique~\cite{21Moravec} and the Vector Field Histogram (VFH) Technique~\cite{22Borenstein,23Borenstein,24Ulrich}. The VFH algorithm is one of the most efficient local path planning algorithms. It generates a two dimensional cartesian histogram grid from the robot's ranging sensors and utilizes the same to create a one dimensional polar grid around the immediate position of the robot. Contiguous sectors with a polar obstacle density below threshold are selected based on the proximity to target direction. This is then utilized by the robot to change steering towards the selected candidate direction. The processed swathe of cells around the arc generated by the swarm algorithm is then utilized to compute a suitable local path.

The VFH algorithm utilizes data from distance ranging sensors like laser or sonar sensors to perform obstacle avoidance. A robot moving across cross-country terrain will have to take into account vagaries of the terrain soil composition.  Therefore, there arises a need to embed additional sensors that will sample the soil in a particular region to determine whether it is suitable for robot traversibility.  The digital cone penetrometer ~\cite{25Johnson,26Perumpral} has been developed primarily to enable rapid assessment of the in-situ strength of the soil. The instrument is used to manually grade the soil and categorize it based on its attributes. The data obtained is matched with the attributes of the various soil types and the region is categorized accordingly.  This soil information is concurrently used along with the obstacle avoidance by the VFH algorithm to aid local path planning. 

\section{Implementation Details and Simulation Results}
The algorithm discussed has been implemented on the open source Player-Stage-Gazebo platform~\cite{27Gerkey}. A simulated  pioneer2DX robot equipped with SICK LMS 200 laser sensor is utilized to test the algorithm on a terrain generated on the Gazebo 3D simulation tool. DEM data from the United States Geographic Services (USGS) was obtained for a region for which free soil information was available in the GRASS GIS project. This DEM data was converted into a height map which was used as input to Gazebo. The effectiveness of the VFH driver in short term path planning was tested out by introducing new obstacles in the form of simulated crates whose presence was not communicated to the grid data fed to the swarm algorithm. A digital cone penetrometer has been simulated as well. Updated soil composition data is fed to the Player client program which parses the soil information into one of the five different categories and suitably moves the robot into safer terrain. 

\begin{figure}[h]
\centering
\includegraphics[width=2in]{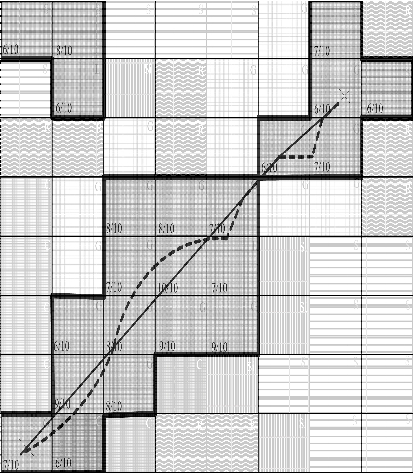}
\caption{Global path and local path.}
\label{fig:soiltypes3}
\end{figure}

\begin{figure}[h]
\centering
\includegraphics[width=3.5in]{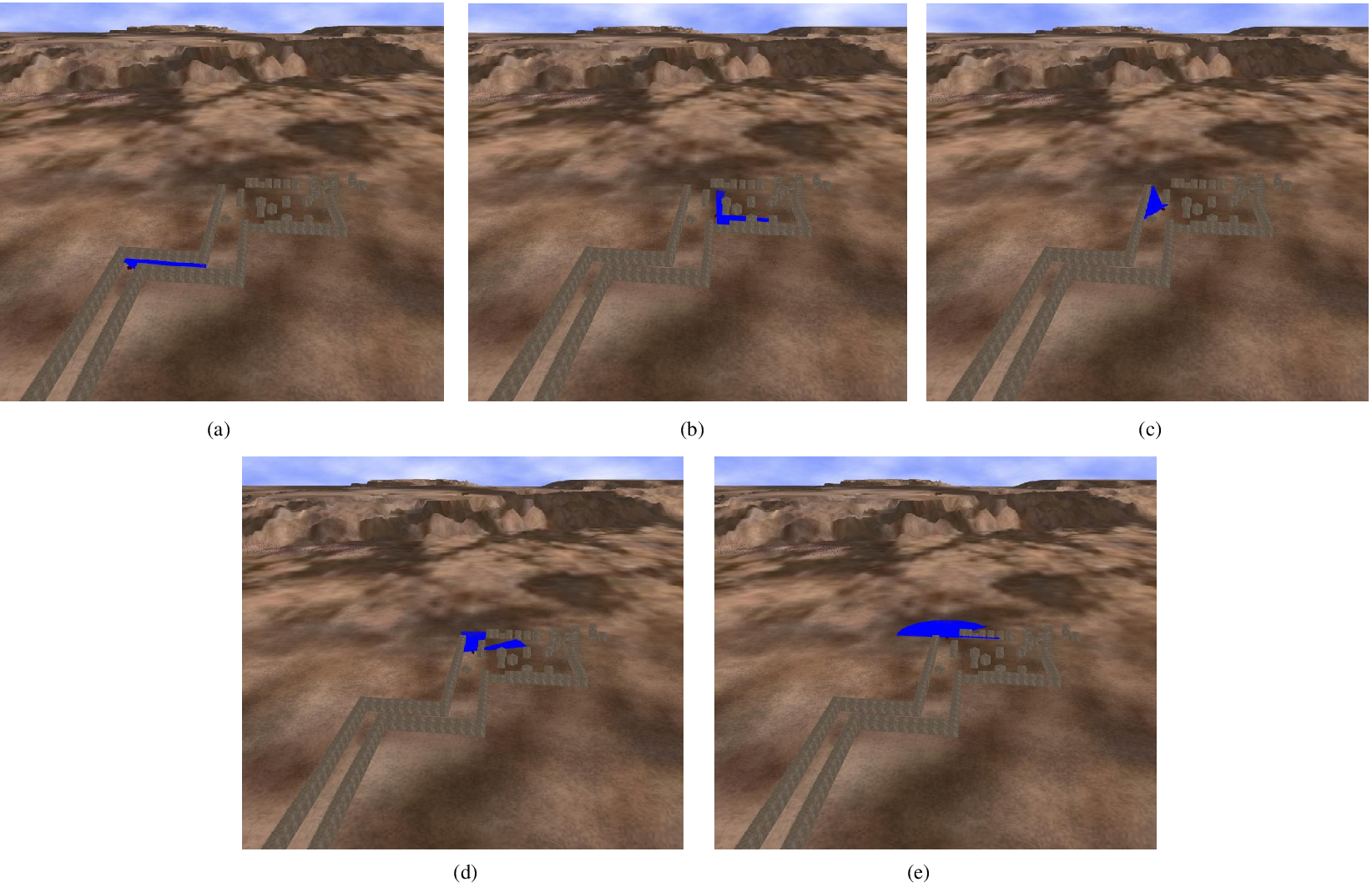}
\caption{Robot~Avoiding~Simulated~Crates~-~Snapshots~of~the~Player~Video.}
\label{fig:all_lcl}
\end{figure}

The entire simulation was run on an Intel Core 2 Duo machine with 256 Mb video RAM and 2 GB RAM. Although the swarm algorithm is designed to run with several agents, the number of agents were limited to 10 because Gazebo's processing power requirements placed constraints on the amount of memory available for other processes to run. Sample soil data is included in Figure~\ref{fig:soiltypes1}. Figure~\ref{fig:soiltypes2} has the combined rank due to soil goodness and heigh information. The set of cells with a dark outline indicate the nest. Figure~\ref{fig:soiltypes3} has the initial global path marked on it and the local variations consequent to obstacle avoidance (shown as a dotted line). Snapshots of the Player video are given in Figure~\ref{fig:all_lcl}.

\section*{Acknowledgment}
The authors would like to thank Director, CAIR for supporting this work.




\begin{thebibliography}{99}
\bibitem{1Laumond}
J.~ Laumond, ``Robot Motion Planning and Control'', {\em Lecture Notes in Control and Information Sciences}, 229, Springer-Verlag New York, Inc., Secaucus, NJ, pp. 305-343, 1998.

\bibitem{2Canny}
J.F. ~Canny, ``The Complexity of Robot Path Planning'', MIT Press, Cambridge, USA, 1988.

\bibitem{3Stappen}
F. ~van ~der ~Stappen, D. ~Halperin, and M.H. ~Overmars, ``The complexity of the free space for a robot moving amidst fat obstacles'', {\em Journal of Computational Geometry: Theory and Applications}, Vol. 3, pp. 353-373, 1993.

\bibitem{4Stappen}
F.~ van ~der ~Stappen, ``Motion Planning amidst Fat Obstacles'', {\em Ph.D. Thesis}, Department of Computer Science, Utrecht University, Utrecht, Netherlands, 1994.

\bibitem{5Schwartz}
J.T. ~Schwartz and M. ~Sharir, ``Efficient motion planning algorithms in environments of bounded local complexity'', {\em Report 164}, Department of Computer Science, Courant Institute of Mathematical Sciences, New York Univ., New york, 1985.

\bibitem{6Svetska}
P. ~Svestka and M.H. ~Overmars, ``Motion planning for car-like robots using a probabilistic learning approach'', {\em International Journal of Robotics Research}, Vol. 16, pp. 119-143, 1995.

\bibitem{7Barraquand}
J. ~Barraquand and J.C. ~Latomber, ``Robot motion planning: A distributed representation approach'', {\em International Journal of Robotics Research}, Vol. 10, pp. 628-649, 1991.

\bibitem{8Bonabeau}
E. ~Bonabeau, M. ~Dorigo and G. ~Theraulaz, ``Swarm Intelligence: From Natural to Artificial Systems'', Oxford University Press, 1999.

\bibitem{9Heero}
K. ~Heero, ``Path planning and learning strategies for mobile robots in dynamic partially unknown environments'', {\em Ph.D. Thesis}, Department of Computer Science, Tartu University, 2006.

\bibitem{10Haigh}
K.Z.~Haigh and M.M.~Veloso, ``Planning, Execution and Learning in a Robotic Agent'', {\em Proceedings of the 4th International Conference on Artificial Intelligence Planning Systems, (AIPS)}, pp. 120-127, 1998.

\bibitem{11Stentz}
A. ~Stentz, ``Optimal and Efficient Path Planning for Partially-Known Environments'', {\em Proceedings of IEEE International Conference on Robotics and Automation}, San Diego, CA, USA, pp. 3310-3317, 1994.

\bibitem{12Stentz}
A. ~Stentz, ``The Focussed D* Algorithm for Real-Time Replanning'', {\em Proceedings of the International Joint Conference on Artificial Intelligence}, Montreal, Quebec, Canada, pp. 1652-1659, 1995.

\bibitem{13Choset}
H.~Choset, K.M.~Lynch, S.~Hutchinson, G.~Kantor, W.~Burgard, L.E.~Kavraki and S.~Thrun, ``Principles of Robot Motion'', The MIT Press, Boston, USA, 2005.

\bibitem{14Hughes}
Ken~Hughes, Alade~Tokuta, N ~Ranganathan, ``Trulla : an algorithm for path planning among weighted regions by localized propagations'', {\em Proceedings of IEEE/RSJ International Conference on Intelligent Robots and Systems}, Raleigh, NC, pp. 469-475, 1992.

\bibitem{15Jahanbin}
M.R. ~Jahanbin and F. ~Fallside, ``Path planning using a wave simulation technique in the configuration space'', {\em Artificial Intelligence in Engineering: Robotics and Processes}, (ed) J.S. Gero, Elsevier Science Publishing, pp 121-139.

\bibitem{16Wilson}
J.P. ~Wilson and J.C. ~Gallant, ``Chapter 1'', {\em In J.P. Wilson and J.C. Gallant (eds.): Terrain Analysis: Principles and Applications}, Wiley, New York, pp. 1-27, 2000.

\bibitem{17Rabus}
Rabus, ~B., M. ~Eineder, A. ~Roth and R. ~Bamler, ``The Shuttle Radar Topography Mission (SRTM)- a new class of digital elevation models acquired by spaceborne radar'', {\em ISPRS Journal of Photogrammetry and Remote Sensing}, Vol. 57, pp. 241-262, 2003.

\bibitem{18usgs}
www.usgs.gov

\bibitem{19Sharma}
A. ~Sharma, V. ~Vyas and D. ~Deodhare, ``An Algorithm for Site Selection in GIS based on Swarm Intelligence'', {\em Proceedings of the IEEE Congress on Evolutionary Computation}, pp. 1020-1027, 2006.

\bibitem{20Kuc}
R. ~Kuc and B. ~Barshan, ``Navigating Vehicles Through an Unstructured Environment With Sonar'', {\em Proceedings of the IEEE International Conference on Robotics and Automation}, Scottsdale, Arizona, pp. 1422-1426, 1989. 

\bibitem{21Moravec}
H.P. ~Moravec, ``Sensor Fusion in Certainty Grids for Mobile Robots'', {\em AI Magazine}, Summer 1988, pp. 61-74.

\bibitem{22Borenstein}
J. ~Borenstein and Y.~Koren, ``The Vector Field Histogram - Fast Obstacle Avoidance for Mobile Robots'', {\em IEEE Journal of Robotics and Automation}, Vol. 3, pp. 278-288, 1991.

\bibitem{23Borenstein}
J. ~Borenstein and Y. ~Koren, ``Real-time Obstacle Avoidance for Fast Mobile Robots in Cluttered Environments'', {\em IEEE Transactions on Systems, Man, and Cybernetics}, Vol. 19, pp. 1179-1187, 1989.

\bibitem{24Ulrich}
I. ~Ulrich and J. ~Borenstein, ``VFH: local obstacle avoidance with look-ahead verification'', {\em Proceedings of the IEEE International Conference on Robotics and Automation, (ICRA)}, San Francisco, CA, USA, Vol. 3, pp. 2505-2511, 2000.

\bibitem{25Johnson}
B.A. ~Johnson and R.M. ~Beard, ``A Lightweight 12-m Cone Penetrometer'', {\em Strength Testing of Marine Sediments:  Laboratory and In-situ Measurements, ASTM STP 883}, R.C.  Chaney and K.R. Demars, Eds., American Society for Testing and Materials, Philadelphia, pp. 125-139, 1985.

\bibitem{26Perumpral}
J.V. ~Perumpral, ``Cone penetrometer applications: A Review'', {\em Transactions of the American Society of  Agricultural Engineers, ASAE}, Beltsville, USA, Vol. 30, pp. 939-944, 1987.

\bibitem{27Gerkey}
B. ~Gerkey, R. ~Vaughan and A. ~Howard, ``The Player/Stage project: Tools for Multi-robot and Distributed Sensor Systems'', {\em Proceedings of the 11th International Conference on Advanced Robotics, (ICAR)}, pp. 317-323, 2003.

\end{thebibliography}
%

\textheight 9.5in
\end{document}